\documentclass[runningheads]{llncs}
\usepackage{graphicx}
\usepackage{comment}
\usepackage{amsmath,amssymb} 
\usepackage{color}
\usepackage{epsfig}
\usepackage{graphicx}
\usepackage{amsmath}
\usepackage{amssymb}
\usepackage{multirow}
\usepackage{subfigure}
\usepackage{emptypage}
\usepackage{verbatimbox}

\usepackage[colorlinks,linkcolor=red]{hyperref}
\usepackage{footmisc}
\usepackage[misc]{ifsym}
\usepackage{cite}

\begin{document}
	\pagestyle{headings}
	\mainmatter
	\def\ECCVSubNumber{4495}  
	
	\title{Consensus-Aware Visual-Semantic Embedding for Image-Text Matching} 
	
	
	\titlerunning{Consensus-Aware Visual-Semantic Embedding}
	
	\author{Haoran Wang\inst{1}\thanks{Work done while Haoran Wang was a Research Intern with Tencent AI Lab.}$^\dagger$\and
		Ying Zhang\inst{2}\thanks{indicates equal contribution.}\and
		Zhong Ji\inst{1}\thanks{Corresponding author} \and
		Yanwei Pang\inst{1} \and
		Lin Ma\inst{2}}
	
	%
	\authorrunning{H. Wang et al.}
	\institute{School of Electrical and Information Engineering, Tianjin University, Tianjin, China \\
		\email{\{haoranwang,Jizhong,pyw\}@tju.edu.cn} \and 
		Tencent AI Lab, Shenzhen, China \\
		\email{yinggzhang@tencent.com\quad forest.linma@gmail.com}}
	
	\maketitle
	
	\begin{abstract}
		Image-text matching plays a central role in bridging vision and language. Most existing approaches only rely on the image-text instance pair to learn their representations, thereby exploiting their matching relationships and making the corresponding alignments. Such approaches only exploit the superficial associations contained in the instance pairwise data, with no consideration of any external commonsense knowledge, which may hinder their capabilities to reason the higher-level relationships between image and text. In this paper, we propose a Consensus-aware Visual-Semantic Embedding (CVSE) model to incorporate the consensus information, namely the commonsense knowledge shared {between both modalities, into image-text matching.} Specifically, the consensus information is exploited by computing the statistical co-occurrence correlations between the semantic concepts from the image captioning corpus and deploying the constructed concept correlation graph to yield the consensus-aware concept (CAC) representations. Afterwards, CVSE learns the associations and alignments between image and text based on the exploited consensus as well as the instance-level representations for both modalities. Extensive experiments conducted on two public datasets verify that the exploited consensus makes significant contributions to constructing more meaningful visual-semantic embeddings, with the superior performances over the state-of-the-art approaches on the bidirectional image and text retrieval task. Our code of this paper is available at: \url{https://github.com/BruceW91/CVSE}.
		
		
	\end{abstract}
	
	\section{Introduction}
	
	Vision and language understanding plays a fundamental role for human to perceive the real world, which has recently made tremendous progresses thanks to the rapid development of deep learning. To delve into multi-modal data comprehending, this paper focuses on addressing the problem of {image-text matching}~\cite{ma2020matching}, which benefits a series of downstream applications, such as visual question answering~\cite{2015VQA,ma2016learning}, visual grounding~\cite{2017Plummer,chen2018temporally,yuan2019semantic}, visual captioning~\cite{wang2018bidirectional,wang2018reconstruction,wang2020reconstruct}, and scene graph generation~\cite{2018KAC}. Specifically, it aims to retrieve the texts (images) that describe the most relevant contents for a given image (text) query. Although thrilling progresses have been made, this task is still challenging
	due to the semantic discrepancy between image and text, which separately resides in heterogeneous representation spaces.

	\begin{figure}[t]
		\begin{center}
			\includegraphics[width=0.9\linewidth,height=6.45cm]{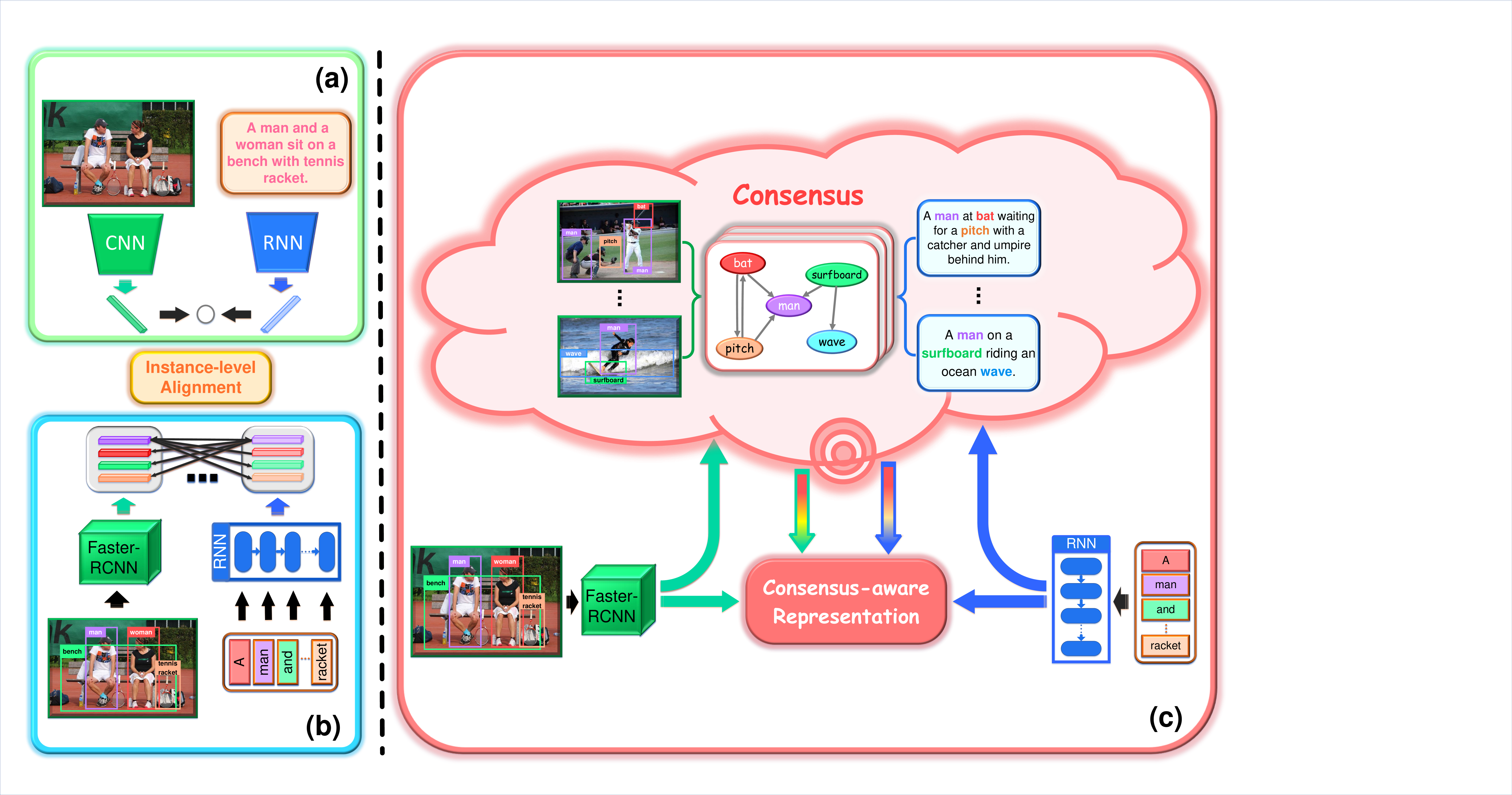}
		\end{center}
		\caption{The conceptual comparison between our proposed consensus-aware visual-semantic embedding (CVSE) approach and existing instance-level alignment based approaches. (a) instance-level alignment based on image and text global representation; (b) instance-level alignment exploiting the complicated fragment-level image-text matching; (c) our proposed CVSE approach.}
		\label{fig.1}
		\label{fig:long}
		\label{fig:onecol}
		\vspace{-0.3cm}
	\end{figure}

	To tackle this problem, the current mainstream solution is to project the image and text into a unified joint embedding space. As shown in Figure \ref{fig.1}~(a), a surge of methods \cite{2014UVSE,2015mrnn,2016Wang,2018VSE++} employ the deep neural networks to extract the global representations of both images and texts, based on which their similarities are measured. However, these approaches failed to explore the relationships between image objects and sentence segments, leading to limited matching accuracy. Another thread of work \cite{2015DVSA,2018SCAN} performs the fragment-level matching and aggregates their similarities to measure their relevance, as shown in Figure \ref{fig.1}~(b). Although complicated cross-modal correlations can be characterized, yielding satisfactory bidirectional image-text retrieval results, these existing approaches only rely on employing the image-text instance pair to perform cross-modal retrieval, which we name as instance-level alignment in this paper.
	
	For human beings, besides the image-text instance pair, we have the capability to leverage our commonsense knowledge, expressed by the fundamental semantic concepts as well as their associations, to represent and align both images and texts. Take one sentence  ``\texttt{A man on a surfboard riding on a ocean wave}'' along with its semantically-related image, shown in Figure \ref{fig.1}~(c), as an example. When ``\texttt{surfboard}'' appears, the word ``\texttt{wave}'' will incline to appear with a high probability in both image and text. As such, the co-occurrence of ``\texttt{surfboard}'' and ``\texttt{wave}'' as well as other co-occurred concepts, constitute the commonsense knowledge, which we refer to as \textit{consensus}. However, such consensus information has not been studied and exploited for the image-text matching task. In this paper, motivated by this cognition ability of human beings, we propose to incorporate the {consensus} to learn visual-semantic embedding for image-text matching. In particular, we not only mine the cross-modal relationships between the image-text instance pairs, but also exploit the consensus from large-scale external knowledge to represent and align both modalities for further visual-textual similarity reasoning.
	
	In this paper, we propose one Consensus-aware Visual-Semantic Embedding (CVSE) architecture for image-text matching, as depicted in Figure \ref{fig.1}~(c). Specifically, we first make the consensus exploitation by computing statistical co-occurrence correlations between the semantic concepts from the image captioning corpus and constructing the concept correlation graph to learn the consensus-aware concept (CAC) representations. Afterwards, based on the learned CAC representations, both images and texts can be represented  at the consensus level. Finally, the consensus-aware representation learning integrates the instance-level and consensus-level representations together, which thereby serves to make the cross-modal alignment. Experiment results on public datasets demonstrate that the proposed CVSE model is capable of learning discriminative representations for image-text matching, and thereby boost the bidirectional image and sentence retrieval performances.
	Our contributions lie in three-fold.
	\begin{itemize}
		\item We make the first attempt to exploit the consensus information for image-text matching. As a departure from existing instance-level alignment based methods, our model leverages one external corpus to learn consensus-aware concept representations expressing the commonsense knowledge for further strengthening the semantic relationships between image and text.
		
		\item We propose a novel Consensus-aware Visual-Semantic Embedding (CVSE) model that unifies the representations of both modalities at the consensus level. And the consensus-aware concept representations are learned with one graph convolutional network, which captures the relationship between semantic concepts for more discriminative embedding learning.
		
		\item The extensive experimental results on two benchmark datasets demonstrate that our approach not only outperforms state-of-the-art methods for traditional image-text retrieval, but also exhibits superior generalization ability for cross-domain transferring.
	\end{itemize}

	\section{Related Work}
	
	\subsection{Knowledge Based Deep Learning}
	
	There has been growing interest in incorporating external knowledge to improve the data-driven neural network. For example, knowledge representation has been employed for image classification \cite{marino2016more} and object recognition \cite{deng2014large}. In the community of vision-language understanding,it has been explored in several contexts, including VQA \cite{wang2018fvqa} and scene graph generation \cite{gu2019scene}. In contrast, our CVSE leverages consensus knowledge to generate homogeneous high-level cross-modal representations and achieves visual-semantic alignment.

	\subsection{Image-Text Matching}
	
	Recently, there have been a rich line of studies proposed for addressing the problem of image-text matching. They mostly deploy the two-branch deep architecture to obtain the global \cite{2014UVSE,2015mrnn,2015mcnn,2016Wang,2018VSE++,ma2020matching} or local \cite{2014DFE,2015DVSA,2018SCAN} representations and align both modalities in the joint semantic space. Mao \emph{et al.} \cite{2015mrnn} adopted CNN and Recurrent Neural Network (RNN) to represent images and texts, followed by employing bidirectional triplet ranking loss to learn a joint visual-semantic embedding space. For fragment-level alignment, Karpathy \emph{et al.} \cite{2015DVSA} measured global cross-modal similarity by accumulating local ones among all region-words pairs. Moreover, several attention-based methods \cite{2017DAN,2018SCAN,Wang2019SaliencyGuidedAN,2019PVSE} have been introduced to capture more fine-grained cross-modal interactions. To sum up, they mostly adhere to model the superficial statistical associations at instance level, whilst the lack of structured commonsense knowledge impairs their reasoning and inference capabilities for multi-modal data.
	
	In contrast to previous studies, our CVSE incorporates the commonsense knowledge into the consensus-aware representations, thereby extracting the high-level semantics shared between image and text. The most relevant existing work to ours  is \cite{shi2019knowledge}, which enhances image representation by employing image scene graph as external knowledge to expand the visual concepts.Unlike \cite{shi2019knowledge}, our CVSE is capable of exploiting the learned consensus-aware concept representations to uniformly represent and align both modalities at the consensus level. Doing so allows us to measure cross-modal similarity via disentangling higher-level semantics for both image and text, which further improves its interpretability.

	\section{Consensus-Aware Visual-Semantic Embedding}
	
\begin{figure*}[!t]
		\centering
		{
			\includegraphics[height=6.5cm,width=0.88\linewidth,]{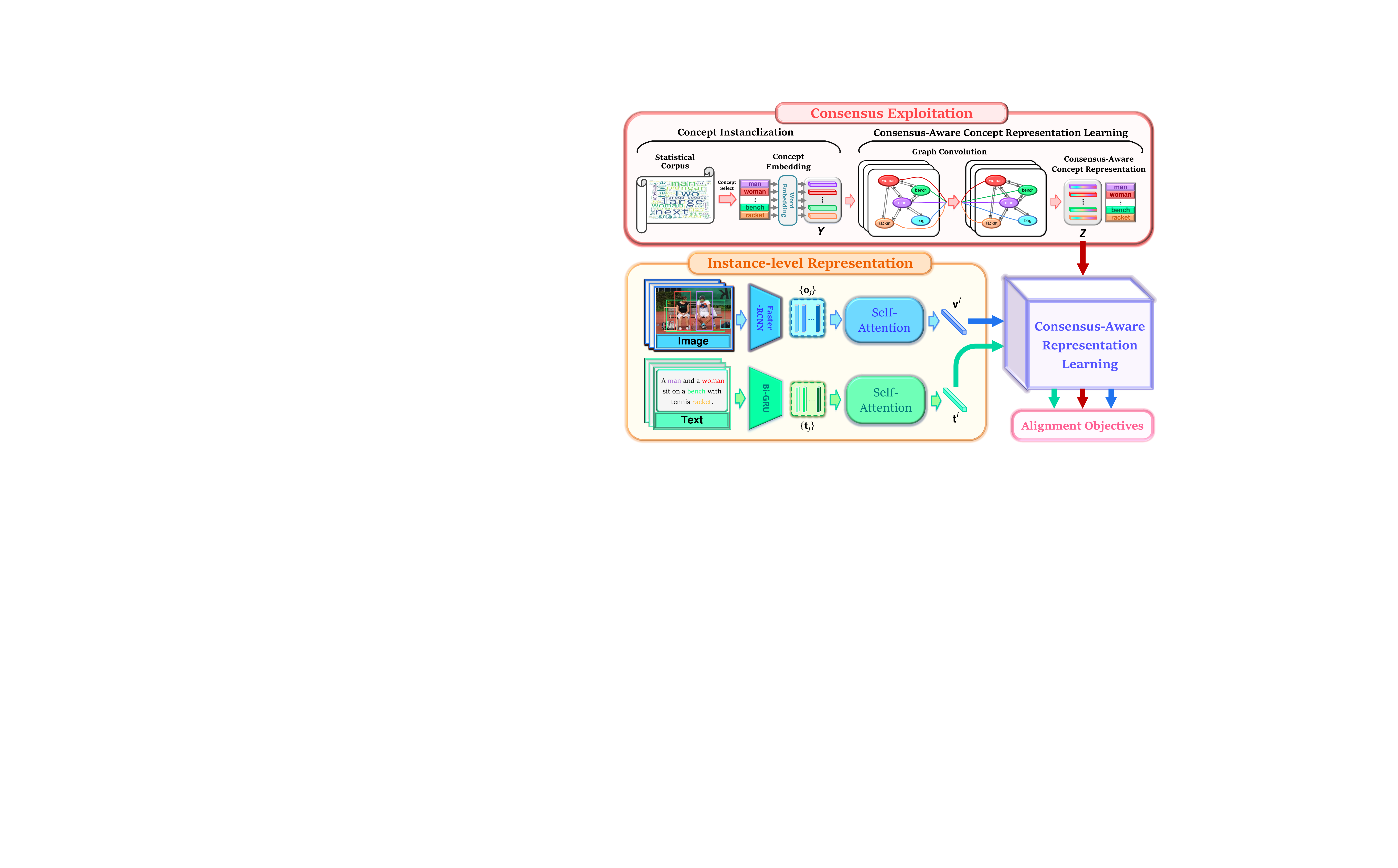}
		}
		\caption{The proposed CVSE model for image-text matching. Taking the fragment-level features of both modalities as input, it not only adopts the dual self-attention to generate the instance-level representations $\mathbf{v}^{I}$ and $\mathbf{t}^{I}$, but also leverages the consensus exploitation module to learn the consensus-level representations.}
		\label{fig.2}	
		\vspace{-0.3cm}	
	\end{figure*}

	In this section, we elaborate on our Consensus-aware Visual-Semantic Embedding (CVSE) architecture for image-text matching (see Figure \ref{fig.2}). Different from instance-level representation based approaches, we first introduce a novel Consensus Exploitation module that leverages commonsense knowledge to capture the semantic associations among concepts. Then, we illustrate how to employ the Consensus Exploitation module to generate the consensus-level representation and combine it with the instance-level representation to represent both modalities. Lastly, the alignment objectives and inference method are represented.

	\subsection{Exploit Consensus Knowledge to Enhance Concept Representations}
		
	As aforementioned, capturing the intrinsic associations among the concepts, which serves as the commonsense knowledge in human reasoning, can analogously supply high-level semantics for more accurate image-text matching. To achieve this, we construct a Consensus Exploitation (CE) module (see Figure \ref{fig.2}), which adopts graph convolution to propagate the semantic correlations among various concepts based on a correlation graph preserving their inter dependencies, which contributes to injecting more commonsense knowledge into the concept representation learning. It involves three key steps: (1) Concept instantiation, (2) Concept correlation graph building and, (3) Consensus-aware concept representation learning. The concrete details will be presented in the following.

	\subsubsection{Concept Instantiation.}
	
	We rely on the image captioning corpus of the natural sentences to exploit the commonsense knowledge, which is represented as the semantic concepts and their correlations. Specifically, all the words in the corpus can serves as the candidate of semantic concepts. Due to the large scale of word vocabulary and the existence of some meaningless words, we follow \cite{fang2015captions,2018SCO} to remove the rarely appeared words from the word vocabulary. In particular, we select the words with top-$q$ appearing frequencies in the concept vocabulary, which are roughly categorized into three types, \textit{i.e.}, \textit{Object}, \textit{Motion}, and \textit{Property}. For more detailed division principle, we refer readers to \cite{hou2019joint}. Moreover, according to the statistical frequency of the concepts with same type over the whole dataset, we restrict the ratio of the concepts with type of (\textit{Object}, \textit{Motion}, \textit{Property}) to be (7:2:1). After that, we employ the glove \cite{pennington2014glove} technique to instantialize these selected concepts, which is denoted as $\mathbf{Y}$.

	\subsubsection{Concept Correlation Graph Building.}
	
	With the instantiated concepts, their co-occurrence relationship are examined to build one correlation graph and thereby exploit the commonsense knowledge. To be more specific, we construct a conditional probability matrix $\mathbf{P}$ to model the correlation between different concepts, with each element $\mathbf{P}_{ij}$ denoting the appearance probability of concept $C_{i}$ when concept $C_{j}$ appears:
	
	\begin{equation}
	\label{eq:pij}
	\begin{aligned}
	\begin{split}
	& \mathbf{P}_{ij} = \left.{\frac{\mathbf{E}_{ij}} {N_{i}}}\right.
	\end{split}
	\end{aligned}
	\end{equation}
	where $\mathbf{E} \in\mathbb{R}^{q \times q}$  is the concept co-occurrence matrix, $\mathbf{E}_{ij}$ represents the co-occurrence times of $C_{i}$ and $C_{j}$, and $N_{i}$ is the occurrence times of $C_{i}$ in the corpus. It is worth noting that  $\mathbf{P}$ is an asymmetrical matrix, which allows us to capture the reasonable inter dependencies among various concepts rather than simple co-occurrence frequency. 			
		
	Although the matrix $\mathbf{P}$ is able to capture the intrinsic correlation among the concepts, it suffers from several shortages. Firstly, it is produced by adopting the statistics of co-occurrence relationship of semantic concepts from the image captioning corpus, which may deviate from the data distribution of real scenario and further jeopardize its generalization ability. Secondly, the statistical patterns derived from co-occurrence frequency between concepts can be easily affected by the long-tail distribution, leading to biased correlation graph. To alleviate the above issues, we design a novel scale function, dubbed Confidence Scaling (CS) function, to rescale the matrix $\mathbf{P}$:	
		
	\begin{equation}	
	\label{eq:bij}	
	\begin{aligned}	
	\begin{split}
	& \mathbf{B}_{ij} = f_{CS}(\mathbf{P}_{ij})= s^{\mathbf{P}_{ij}-u}-s^{-u},	
	\end{split}	
	\end{aligned}	
	\end{equation}	
	\MakeLowercase{where} $s$ and $u$ are two pre-defined parameters to determine the the amplifying/shrinking rate for rescaling the elements of $\mathbf{P}$. Afterwards, to further prevent the correlation matrix from being over-fitted to the training data and improve its generalization ability, we also follow \cite{chen2019multi} to apply binary operation to the rescaled matrix $\mathbf{B}$:	
	\begin{equation}	
	\label{eq:gij}	
	\mathbf{G}_{ij} = \left\{	
	\begin{aligned}
	0 & , & if \ \; \mathbf{B}_{ij} < \ \epsilon, \\
	1 & , & if \ \; \mathbf{B}_{ij} \ge \  \epsilon,
	\end{aligned}
	\right.
	\end{equation}
	\MakeLowercase{where} $\mathbf{G}$ is the binarized matrix $\mathbf{B}$. $\epsilon$ denotes a threshold parameter filters noisy edges. Such scaling strategy not only assists us to focus on the more reliable co-occurrence relationship among the concepts, but also contributes to depressing the noise contained in the long-tailed data.

	\subsubsection{Consensus-Aware Concept Representation.}
	
	Graph Convolutional Network (GCN) \cite{bruna2013spectral,kipf2016semi} is a multilayer neural network that operates on a graph and update the embedding representations of the nodes via propagating information based on their neighborhoods. Distinct from the conventional convolution operation that are implemented on images with Euclidean structures, GCN can learn the mapping function on graph-structured data. In this section, we employ the multiple stacked GCN layers to learn the concept representations (dubbed CGCN module), which introduces higher order neighborhoods information among the concepts to model their inter dependencies. More formally, given the instantiated concept representations $\mathbf{Y}$ and the concept correlation graph $\mathbf{G}$, the embedding feature of the $l$-th layer is calculated as
	\begin{equation}
	\label{eq:hl}
	\begin{aligned}
	\begin{split}
	& \mathbf{H}^{(l+1)} = \rho (\tilde{\mathbf{A}}\mathbf{H}^{(l)}\mathbf{W}^{(l)})
	\end{split}
	\end{aligned}
	\end{equation}	where  $\mathbf{H}^{(0)}=\mathbf{Y}$,  $\tilde{\mathbf{A}}=\mathbf{D}^{-\frac{1}{2}}\mathbf{G}\mathbf{D}^{-\frac{1}{2}}$ denotes the normalized symmetric matrix and $\mathbf{W}^{l}$ represents the learnable weight matrix.  $\rho$ is a non-linear activation function, \textit{e.g.}, ReLU function \cite{krizhevsky2012imagenet}.
	
	We take output of the last layer from GCN to acquire the final concept representations $\mathbf{Z}\in\mathbb{R}^{q \times d}$ with $\mathbf{z}_{i}$ denoting the generated embedding representation of concept $C_i$, and  $d$ indicating the dimensionality of the joint embedding space.
	Specifically, the $i$-th row vector of matrix $\mathbf{Z}=\left\{ {\mathbf{z}_{1},...,\mathbf{z}_{q}} \right\}$, \textit{i.e.} $\mathbf{z}_{i}$, represents the embedding representation for the $i$-th element of the concept vocabulary. For clarity, we name $\mathbf{Z}$ as consensus-aware concept (CAC) representations, which is capable of exploiting the commonsense knowledge to capture underlying interactions among various semantic concepts.

	\subsection{Consensus-Aware Representation Learning}
	
	In this section, we would like to incorporate the exploited consensus to generate the consensus-aware representation of image and text.
	
	\subsubsection{Instance-level Image and Text Representations.}
	As aforementioned, conventional image-text matching only rely on the individual image/text instance to yield the correponding representations for matching, as illustrated in Figure~\ref{fig.2}. Specifically, given an input image, we utilize a pre-trained Faster-RCNN \cite{2015FasterRcnn,2018Bottomup} followed by a fully-connected (FC) layer to represent it by $M$ region-level visual features $\mathbf{O}=\left\{ {\mathbf{o}_{1},...,\mathbf{o}_{M}} \right\}$, whose elements are all $F$-dimensional vector. Given a sentence with $L$ words, the word embedding is sequentially fed into a bi-directional GRU \cite{schuster1997bidirectional}. After that, we can obtain the word-level textual features $\left\{ {\mathbf{t}_{1},...,\mathbf{t}_{L}} \right\}$ by performing mean pooling to aggregate the forward and backward hidden state vectors at each time step.
	
	Afterwards, the self-attention mechanism \cite{2017AAN} is used to concentrate on the informative portion of the fragment-level features to enhance latent embeddings for both modalities. Note that here we only describe the attention generation procedure of the visual branch, as it goes the same for the textual one. The region-level visual features $\left\{ {\mathbf{o}_{1},...,\mathbf{o}_{M}} \right\}$ is used as the key and value items, while the global visual feature vector $\bar{\mathbf{O}} = \frac{1}{M}\sum\nolimits_{m = 1}^M {{\mathbf{o}_m}}$ is adopted as the query item for the attention strategy. As such, the self-attention mechanism refines the instance-level visual representation as $\mathbf{v}^I$. With the same process on the word-level textual features $\left\{ {\mathbf{t}_{1},...,\mathbf{t}_{L}} \right\}$, the instance-level textual representation is refined as $\mathbf{t}^I$.

	\begin{figure*}[!t]
		\centering
		{
			\includegraphics[height=3.8cm,width=0.87\linewidth,]{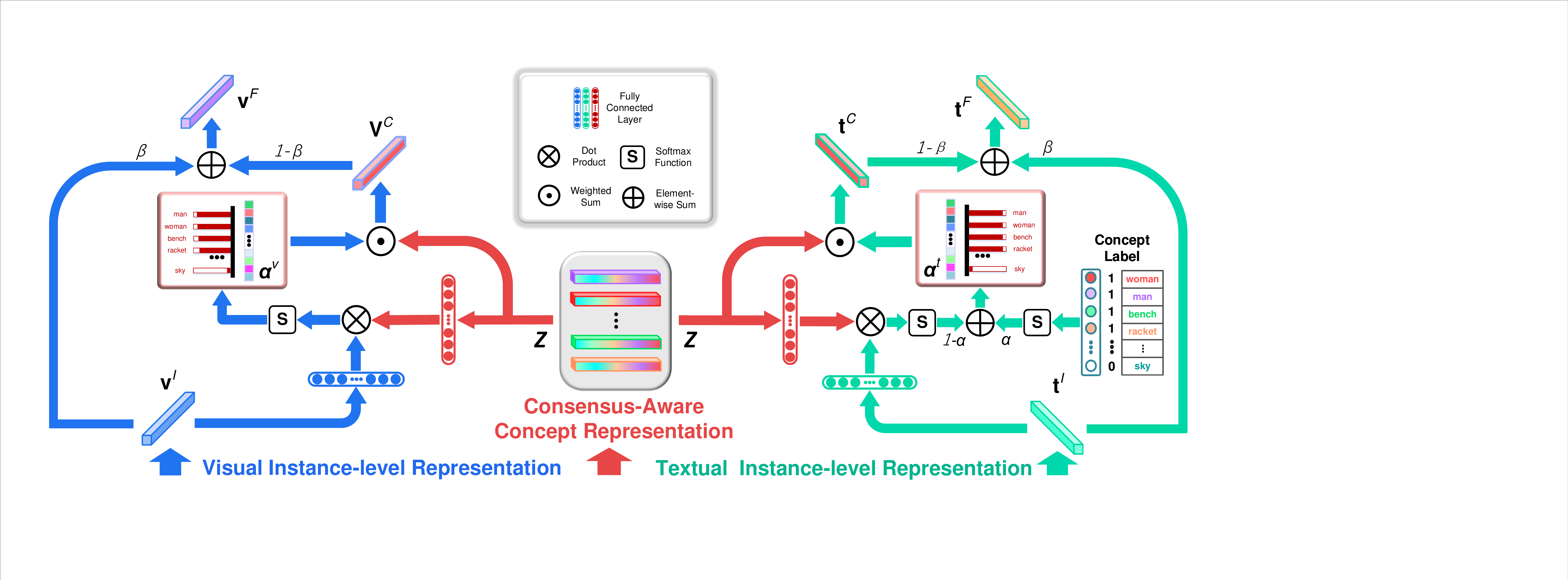}
		}
		\setlength{\abovecaptionskip}{+0.25cm}
		\caption{Illustration of the consensus-level representation learning and the fusion between it and instance-level representation.}
		\label{fig.3}
		\vspace{-0.3cm}
	\end{figure*}

	\subsubsection{Consensus-level Image and Text Representations.}
	
	In order to incorporate the exploited consensus, as shown in Figure~\ref{fig.3}, we take the instance-level visual and textual representations ($\mathbf{v}^I$ and $\mathbf{t}^I$) as input to query from the CAC representations. The generated significance scores for different semantic concepts allow us to uniformly utilize the linear combination of the CAC representations to represent both modalities. Mathematically, the visual consensus-level representation $\mathbf{v}^C$ can be calculated as follows:
	\begin{equation}
	\label{eq:vc}
	\begin{aligned}
	\begin{split}
	& \mathbf{a}_{i}^v = \left.{\frac
		{exp (\lambda \mathbf{v}^{I} {\mathbf{W}}^{v} \mathbf{z}_{i}^\mathsf{T})}
		{\sum\nolimits_{i = 1}^q exp (\lambda \mathbf{v}^{I} {\mathbf{W}}^{v} \mathbf{z}_{i}^\mathsf{T})}}\right., \\
	& \mathbf{v}^C = \sum\nolimits_{i = 1}^q {\mathbf{a}_{i}^v} \cdot \mathbf{z}_{i},
	\end{split}
	\end{aligned}
	\end{equation}
	\MakeLowercase{where} ${\mathbf{W}}^{v} \in\mathbb{R}^{d \times d}$ is the learnable parameter matrix, $\mathbf{a}_{i}^v$ denotes the significance score corresponding to the semantic concept $\mathbf{z}_i$, and $\lambda$ controls the smoothness of the softmax function.
	
	For the text, due to the semantic concepts are instantiated from the textual statistics, we can annotate any given image-text pair via employing a set of concepts that appears in its corresponding descriptions. Formally, we refer to this multi-label tagging as concept label ${\mathbf{L}}^{t}\in\mathbb{R}^{q \times 1}$. Considering the consensus knowledge is explored from the textual statistics, we argue that it's reasonable to leverage the concept label as prior information to guide the consensus-level representation learning and alignment.  Specifically, we compute the predicted concept scores $\mathbf{a}_{i}^t$ and consensus-level representation $\mathbf{t}^C$ as follows:	
	
	\begin{equation}	
	\label{eq:tc}	
	\begin{aligned}		
	\begin{split}	
	& \mathbf{a}_{j}^t = \alpha \left.{\frac
		{exp (\lambda \mathbf{L}_{j}^{t})}
		{\sum\nolimits_{j = 1}^q exp (\lambda \mathbf{L}_{j}^{t})}}\right.
	+
	(1-\alpha) \left.{\frac
		{exp (\lambda \mathbf{t}^{I} {\mathbf{W}}^{t} \mathbf{z}_{j}^\mathsf{T})}
		{\sum\nolimits_{j = 1}^q exp (\lambda \mathbf{t}^{I} {\mathbf{W}}^{t} \mathbf{z}_{j}^\mathsf{T})}}\right.,
	\\
	& \mathbf{t}^C = \sum\nolimits_{j = 1}^q {\mathbf{a}_{j}^t} \cdot \mathbf{z}_{j},
	\end{split}
	\end{aligned}
	\end{equation}
	where ${\mathbf{W}}^{t}\in\mathbb{R}^{d \times d}$ denotes the learnable parameter matrix. $\alpha\in[0,1]$ controls the proportion of the concept label to generate the textual predicted concept scores $\mathbf{a}_{j}^t$. We empirically find that incorporating the concept label into the textual consensus-level representation learning can significantly boost the performances.

	\subsubsection{Fusing Consensus-level and Instance-level Representations.}
	We integrate the instance-level representations  $\mathbf{v}^I$($\mathbf{t}^I$) and consensus-level representation $\mathbf{v}^C$($\mathbf{t}^C$) to  comprehensively characterizing the semantic meanings of the visual and textual modalities. Empirically, we find that the simple weighted sum operation can achieve satisfactory results, which is defined as:
	\begin{equation}
	\label{eq:tfvf}
	\begin{aligned}	
	\begin{split}
	& \mathbf{v}^F = \beta \mathbf{v}^I + (1-\beta) \mathbf{v}^C, \\
	& \mathbf{t}^F = \beta \mathbf{t}^I + (1-\beta) \mathbf{t}^C,
	\end{split}
	\end{aligned}
	\end{equation}
	\MakeLowercase{where} $\beta$ is a tuning parameter controlling the ratio of two types of representations. And $\mathbf{v}^F$ and $\mathbf{t}^F$ respectively denote the combined visual and textual representations, dubbed consensus-aware representations.

	\subsection{Training and Inference}
	
	\subsubsection{Training.} During the training, we deploy the widely adopted bidirectional triplet ranking loss \cite{frome2013devise,2014UVSE,2018VSE++} to align the image and text:
	\begin{equation}
	\label{eq:lrank}
	\begin{aligned}
	\begin{array}{l}
	\mathcal{L}_{rank}(\mathbf{v},\mathbf{t}) = \sum\limits_{(\mathbf{v},\mathbf{t})} \{  \max [0, {\gamma} - s(\mathbf{v},\mathbf{t}) + s({\mathbf{v}},\mathbf{t}^-)] \\
	\begin{array}{*{10}{c}}
	{\begin{array}{*{10}{c}}
		{\begin{array}{*{10}{c}}
			{}&&&&&&&&&{}
			\end{array}}&&&&&&&&&{}
		\end{array}}&{}
	\end{array}
	+ \max [0,{\gamma} - s(\mathbf{t},\mathbf{v}) + s(\mathbf{t},{\mathbf{v}^ - })]\},
	\end{array}
	\end{aligned}
	\end{equation}
	where $\gamma$ is a predefined margin parameter, $s\left( { \cdot , \cdot } \right)$ denotes cosine distance function. Given the representations for a matched image-text pair $ (\mathbf{v},\mathbf{t}) $, its corresponding negative pairs are denoted as $(\mathbf{t}, \mathbf{v}^{-})$ and $ (\mathbf{v},\mathbf{t}^{-})$, respectively. The bidirectional ranking objectives are imposed on all three types of representations, including instance-level, consensus-level, and consensus-aware representations.
	
	Considering that a matched image-text pair usually contains similar semantic concepts, we impose the Kullback Leibler (KL) divergence on the visual and textual predicted concept scores to further regularize the alignment:  	
	\begin{equation}
	\label{eq:dkl}
	\begin{aligned}	
	\begin{split}
	& \mathcal{D}_{KL}(\mathbf{a}^t\ \| \ \mathbf{a}^v) = \sum\nolimits_{i = 1}^q {\mathbf{a}_{i}^t} \ log (\frac{\mathbf{a}_{i}^t}{\mathbf{a}_{i}^v}),
	\end{split}
	\end{aligned}
	\end{equation}
	
	In summary, the final training objectives of our CVSE model is defined as:
	\begin{equation}
	\label{eq:lall}
	\begin{aligned}
	\begin{split}
	&	\mathcal{L} = {\lambda }_{1} {\mathcal{L}}_{rank}(\mathbf{v^F},\mathbf{t^F}) + {\lambda }_{2} {\mathcal{L}}_{rank}(\mathbf{v^I},\mathbf{t^I}) +
	{\lambda }_{3} {\mathcal{L}}_{rank}(\mathbf{v^C},\mathbf{t^C}) + {\lambda }_{4} \mathcal{D}_{KL},
	\end{split}
	\end{aligned}
	\end{equation}
	where ${\lambda }_{1}, {\lambda }_{2}, {\lambda }_{3}, {\lambda }_{4}$ aim to balance the weight of different loss functions.

	\subsubsection{Inference.} During inference, we only deploy the consensus-aware representations $\mathbf{v}^F$ ($\mathbf{t}^F$) and utilize cosine distance to measure their cross-modal similarity. {Due to we employ the shared concept labels from pairwise sentences for model training, a concept prediction strategy is utilized to narrow the gap between training and inference stage. Specifically, given the consensus-aware visual and textual representations, we predict the relevant concepts for a textual description from two perspectives: 1). Text-to-text similarity. We perform $K$-nearest neighbour (KNN) search according to textual similarity and obtain the index of $k$ most relevant sentences $\mathbf{I}^{k}_{t2t}$. 2). Cross-modal similarity. According to the text-to-image similarity, we first locate an image that is most relevant to the given sentence, following by using KNN search to acquire the index of $k$ nearest sentences $\mathbf{I}^{k}_{i2t}$ by measuring image-to-text similarity. Finally, we merge the $\mathbf{I}^{k}_{t2t}$ and $\mathbf{I}^{k}_{i2t}$ together and employ the union of their own concept labels as predicted concept labels. Moreover, from another view, our concept prediction method can be also taken as a special re-ranking process, which is commonly used in intra-modal \cite{Zhong2017RerankingPR} and cross-modal \cite{Wang2019MatchingIA} retrieval tasks. }

\section{Experiments}

\subsection{Dataset and Settings}

\subsubsection{Datasets.}
{Flickr30k} \cite{2015Flickr30k} is an image-caption dataset containing 31,783 images, with each image annotated with five sentences. Following the protocol of \cite{2015mrnn}, we split the dataset into 29,783 training, 1000 validation, and 1000 test images. We report the performance evaluation of image-text retrieval on 1000 test set.
{MSCOCO} \cite{2014COCO} is another image-caption dataset that contains 123,287 images with each image roughly annotated with five sentence-level descriptions. We follow the public dataset split of \cite{2015DVSA}, including 113,287 training images, 1000 validation images, and 5000 test images. We report the experimental results on 1K test set, which is averaging over 5 folds of 1K set of the full 5K test images.

\subsubsection{Evaluation Metrics.}We employ the widely-used R@K as evaluation metric \cite{2014UVSE,2018VSE++}, which measures the the fraction of queries for which the matched item is found among the top \emph{k} retrieved results. We also report the ``mR'' criterion that averages all six recall rates of R@K, which provides a more comprehensive evaluation to testify the overall performance.

\subsection{Implementation Details}
All our experiments are implemented in PyTorch with one NVIDIA Tesla P40 GPU. For visual representation, the amount of detected regions in each image is $M=36$, and the dimensionality of region vectors is $F=2048$. The dimensionality of word embedding space is set to 300. The dimensionality of joint space $d$ is set to 1024. For the consensus exploitation, we adopt 300-dim GloVe \cite{pennington2014glove} trained on the Wikipedia dataset to initialize the the semantic concepts. The size of the semantic concept vocabulary is $q=300$. And two graph convolution layers are used, with the embedding dimensionality are set to 512 and 1024, respectively. For the correlation matrix $\mathbf{G}$, we set $s=5$ and $u=0.02$ in Eq.~\eqref{eq:bij}, and $\epsilon=0.3$ in Eq.~\eqref{eq:gij}. For image and text representation learning, we set $\lambda=10$ in Eq.~\eqref{eq:vc} and $\alpha=0.35$ in Eq.~\eqref{eq:tc}, respectively. For the training objective, we empirically set $\beta=0.75$ in Eq.~\eqref{eq:tfvf} , $\gamma=0.2$ in Eq.~\eqref{eq:lrank} and ${\lambda }_{1}, {\lambda }_{2}, {\lambda }_{3}, {\lambda }_{4}=3, 5, 1, 2$ in Eq.~\eqref{eq:lall}. For inference, we set $k=3$. Our CVSE model is trained by Adam optimizer \cite{2014Adam} with mini-batch size of 128. The learning rate is set to be 0.0002 for the first 15 epochs and 0.00002 for the next 15 epochs. The dropout is employed with a dropout rate of 0.4. Our code is available \footnote{\url{https://github.com/BruceW91/CVSE}}.

\setlength{\tabcolsep}{4pt}
\begin{table}[t]
	
	\begin{center}
		\caption{Comparisons of experimental results on MSCOCO 1K test set and Flickr30k test set. }
		\label{tab.1}
		\resizebox{0.99\textwidth}{2.2cm}{
			\begin{tabular}{l|ccc|ccc|c|ccc|ccc|c}
				\hline
				\hline
				\multicolumn{1}{c|}{\multirow{3}{*}{Approach}}  & \multicolumn{7}{c|}{ MSCOCO  dataset}                                                                       & \multicolumn{7}{c}{Flickr30k dataset}                                                                                                                                                   \\
				\hline
				& \multicolumn{3}{c|}{Text retrieval}                                       & \multicolumn{3}{c|}{Image Retrieval}                                          & \multicolumn{1}{c|}{\multirow{2}{*}{mR}} & \multicolumn{3}{c|}{Text retrieval}                                       & \multicolumn{3}{c}{Image Retrieval}
				& \multicolumn{1}{|c}{\multirow{2}{*}{mR}}\\
				
				\multicolumn{1}{c}{}                           & \multicolumn{1}{|c}{R@1} & \multicolumn{1}{c}{R@5} & \multicolumn{1}{c|}{R@10} & \multicolumn{1}{c}{R@1} & \multicolumn{1}{c}{R@5} & \multicolumn{1}{c|}{R@10} & \multicolumn{1}{c|}{}   & \multicolumn{1}{c}{R@1} & \multicolumn{1}{c}{R@5} & \multicolumn{1}{c|}{R@10} & \multicolumn{1}{c}{R@1} & \multicolumn{1}{c}{R@5} & \multicolumn{1}{c|}{R@10} & \multicolumn{1}{c}{}   \\
				\hline
				
				DVSA \cite{2015DVSA}         & 38.4                    & 69.9                    & 80.5                     & 27.4                    & 60.2                    & 74.8                     & 39.2                   & 22.2                    & 48.2                    & 61.4                     & 15.2                    & 37.7                    & 50.5                     & 58.5                   \\
				
				m-RNN \cite{2015mrnn}      &	41.0 &	73.0 &	83.5 &	29.0 &	42.2 &	77.0  &	57.6 &	 35.4 & 63.8 & 73.7 & 22.8 & 50.7 & 63.1 & 51.6  \\
				
				DSPE \cite{2016Wang}    & 50.1                    & 79.7                    & 89.2                     & 39.6                    & 75.2                    & 86.9                     & 70.1             		& 40.3                    & 68.9                    & 79.9                     & 29.7                    & 60.1                    & 72.1                     & 58.5      \\
				
				CMPM \cite{2018CMPM}  & 56.1 &	86.3 &	92.9  &	44.6 &	78.8 &	89  & 74.6   & 49.6 &	 76.8 &	86.1  &	37.3 &	65.7 &	75.5  &	65.2  \\
				
				VSE++ \cite{2018VSE++}      & 64.7                    & -                       & 95.9                     & 52.0                    & -                       & 92.0                     & -         & 52.9                    & -                       & 87.2                     & 39.6                    & -                       & 79.5                     & -              \\
				
				PVSE \cite{2019PVSE}   & 69.2                   & 91.6                      & 96.6                     & 55.2                       & 86.5                   & 93.7	& -  & -  & -  & -  & -  & -  & -                \\
				
				SCAN \cite{2018SCAN}   & 72.7  &	94.8  &	\textbf{98.4} &	58.8 &	88.4 &	94.8  &	83.6   &	 67.4 &	90.3 &	\textbf{95.8} &	48.6 &	77.7 &	85.2 &	77.5	\\
				
				CAMP \cite{wang2019camp}	& 72.3 &	94.8 &	98.3 &	58.5 &	87.9 &	95.0  &	84.5   &	 68.1 &	89.7 &	95.2 &	51.5 &	77.1 &	85.3 &	77.8	\\
				
				LIWE \cite{wehrmann2019language}	& 73.2 &	\textbf{95.5} &	98.2 &	57.9 &	88.3 &	94.5  &	 84.6   &	69.6 &	90.3 &	95.6 &	51.2 &	\textbf{80.4} &	87.2 &	79.1	\\
				
				\hline		
				{CVSE}
				& {\textbf{74.8}} &	{95.1} &	{98.3} &	{\textbf{59.9}}
				& {\textbf{89.4}} &	{\textbf{95.2}}  & {\textbf{85.5}}  &	{\textbf{73.5}} &	{\textbf{92.1}} &	{\textbf{95.8}}  &	 {\textbf{52.9}} &	{\textbf{80.4}} &	{\textbf{\textbf{87.8}}}  & {\textbf{80.4}} 			
				
				\\
				\hline
				\hline
			\end{tabular}
		}
	\end{center}
	\vspace{-0.5cm}
\end{table}
\setlength{\tabcolsep}{1.4pt}

\subsection{Comparison to State-of-the-art}

The experimental results on the MSCOCO dataset are shown in Table \ref{tab.1}. From Table \ref{tab.1}, we can observe that our CVSE is obviously superior to the competitors in most evaluation metrics, which yield a result of 74.8\% and 59.9\% on R@1 for text retrieval and image retrieval, respectively. In particular, compared with the second best LIWE method, we achieve absolute boost (2.0\%, 1.1\%, 1.0\%) on (R@1, R@5, R@10) for image retrieval. Moreover, as the most persuasive criteria, the mR metric of our CVSE still markedly exceeds other algorithms. Besides, some methods partially surpassed ours, such as SCAN \cite{2018SCAN} exhaustively aggregating the local similarities over the visual and textual fragments, which leads to slow inference speed. By contrast, our CVSE just employs combined global representations so that substantially speeds up the inference stage. Therefore, considering the balance between effectiveness and efficiency, our CVSE still has distinct advantages over them.

The results on the Flickr30K dataset are presented in Table \ref{tab.1}. It can be seen that our CVSE arrives at 80.4\% on the criteria of ``mR'', which also outperforms all the state-of-the-art methods. Especially for text retrieval, the CVSE model surpasses the previous best method by (3.9\%, 1.8\%, 0.2\%) on (R@1, R@5, R@10), respectively. The above results substantially demonstrate the effectiveness and necessity of exploiting the consensus between both modalities to align the visual and textual representations.

\setlength{\tabcolsep}{4pt}
\begin{table}[t]
	\setlength{\belowcaptionskip}{+0.15cm}
	
	\begin{center}
		\caption{Effect of different configurations of CGCN module on MSCOCO Dataset.}
		\label{tab.2}
		\centering	
		\resizebox{0.99\textwidth}{1.0cm}{	
			\begin{tabular}{l|c|c|c|ccc|ccc}
				\hline \hline
				\multirow{2}{*}{Approaches} &
				\multicolumn{3}{c}{CGCN} &
				\multicolumn{3}{|c}{Text Retrieval} & \multicolumn{3}{|c}{Image Retrieval}  \\
				& Graph Embedding   & CS Function & Concept Label & R@1 & R@5 & R@10 & R@1 & R@5 & R@10 \\
				\hline
				
				{CVSE$_{full}$}	& $\checkmark$ & $\checkmark$ & $\checkmark$
				& {\textbf{74.8}} &	{\textbf{95.1}} &	{\textbf{98.3}} &	{\textbf{59.9}} & {\textbf{89.4}} &	{\textbf{95.2}}	\\
				
				\hline
				{CVSE$_{wo/GE}$} 	& $\quad$ & $\checkmark$ & $\checkmark$ & {71.5}
				& {93.5}	& {97.2}	& {55.1}	& {87.3}	& {92.7}	 \\
				
				{CVSE$_{wo/CS}$} 	& $\checkmark$ & $\quad$ &  $\checkmark$ & {74.5}	& {94.7}	& {97.8}	& {58.8}	& {88.7}	& {94.7}	 \\		
				
				CVSE$_{wo/CL}$	& $\checkmark$ & $\checkmark$ & $\quad$ & 72.5	& 93.5	& 97.7	& 57.2	& 87.4		& 94.1 \\

				\hline \hline
				
			\end{tabular}
		}
	\end{center}
\end{table}

\setlength{\tabcolsep}{4pt}
\begin{table}[!t]
	\begin{center}
		\caption{Effect of different configurations of objective and inference scheme on MSCOCO Dataset.}
		\label{tab.3}
		\centering	
		\resizebox{0.99\textwidth}{0.95cm}{	
			\begin{tabular}{l|c|c|c|c|c|ccc|ccc}
				\hline \hline
				\multirow{2}{*}{Approaches} &
				\multicolumn{2}{c}{Objective} & \multicolumn{3}{|c}{Inference Scheme} &
				\multicolumn{3}{|c}{Text retrieval} & \multicolumn{3}{|c}{Image Retrieval}  \\
				& Separate Constraint & KL & Instance & Consensus & Fused		&   R@1 & R@5 & R@10 & R@1 & R@5 & R@10 \\
				\hline			
				{CVSE$_{wo/SC}$ }	&	& $\checkmark$ 	&	&	& $\checkmark$ 
				& {72.9}	& {94.8}	& {97.7}	& {59.0}	 & {88.8}	& {94.1}	 \\	
				
				{CVSE$_{wo/KL}$ } 	& $\checkmark$ 	&	&	&	& $\checkmark$ & {\textbf{74.4}}	& {\textbf{95.0}}	& {\textbf{97.9}}	 & {\textbf{59.6}}	& {\textbf{89.1}}	& {\textbf{94.7}}	 \\
				
				\hline
				CVSE ($\beta=1$) & $\checkmark$ & $\checkmark$ & $\checkmark$ &	&	& 71.2	& 93.8	 & 97.4		& 54.8	& 87.0	& 92.2	 \\		
				
				{CVSE ($\beta=0$) } & $\checkmark$ & $\checkmark$ &	& $\checkmark$ &  &	{47.6}	& {70.6}	& {82.1}	& {41.7}	& {70.2}	& {80.8}	 \\	
				
				\hline \hline
				
			\end{tabular}
		}
	\end{center}
	\vspace{-0.6cm}
\end{table}

\subsection{Ablation Studies}

In this section, we perform several ablation studies to systematically explore the impacts of different components in our CVSE model. Unless otherwise specified, we validate the performance on the 1K test set of MSCOCO dataset.

\subsubsection{Different Configuration of Consensus Exploitation.} To start with, we explore how the different configurations of consensus exploitation module affects the performance of CVSE model. As shown in Table \ref{tab.2}, it is observed that although we only adopt the glove word embedding as the CAC representations, the model (CVSE$_{wo/GC}$) can still achieve the comparable performance in comparison to the current leading methods. It indicates the semantic information contained in word embedding technique is still capable of providing weak consensus information to benefit image-text matching. Compared to the model (CVSE$_{wo/CS}$) where CS function in Eq.~\eqref{eq:bij} is excluded, the CVSE model can obtain 0.3\% and 1.1\% performance gain on R@1 for text retrieval and image retrieval, respectively. Besides, we also find that if the concept label ${\mathbf{L}}^{t}$ is excluded, the performance of model (CVSE$_{wo/CL}$) evidently drops. We conjecture that this result is attributed to the consensus knowledge is collected from textual statistics, thus the textual prior information contained in concept label substantially contributes to enhancing the textual consensus-level representation so as to achieve more precise cross-modal alignment.

\subsubsection{Different Configurations of Training Objective and Inference Strategy.} We further explore how the different alignment objectives affect our performance. First, as shown in Table \ref{tab.3}, when the separate ranking loss, \textit{i.e.} ${\mathcal{L}}_{rank-I}$ and $ {\mathcal{L}}_{rank-C}$ are both removed, the CVSE$_{wo/SC}$ model performs worse than our CVSE model, which validates the effectiveness of the two terms. Secondly, we find that the CVSE$_{wo/KL}$ produces inferior retrieval results, indicating the importance of ${\mathcal{D}}_{KL}$ for regularizing the distribution discrepancy of predicted concept scores between image and text, which again provides more interpretability that pariwise heterogeneous data should correspond to the approximate semantic concepts. Finally, we explore the relationship between instance-level features and consensus-level features for representing both modalities. Specifically, the CVSE ($\beta=1$) denotes the CVSE model with $\beta=1$ in Eq.~\eqref{eq:tfvf}, which employs instance-level representations alone. Similarly, the CVSE ($\beta=0$) model refers to the CVSE that only adopts the consensus-level representations. Interestingly, we observe that deploying the representations from any single semantic level alone will yields inferior results compared to their combination. It substantially verifies the semantical complementarity between the instance-level and consensus-level representations is critical for achieving significant performance improvements.

\subsection{Further Analysis}

\subsubsection{Consensus Exploitation for Domain Adaptation.} To further verify the capacity of consensus knowledge, we test its generalization ability by conducting cross-dataset experiments, which was seldom investigated in previous studies whilst meaningful for evaluating the cross-modal retrieval performance in real scenario. Specifically, we conduct the experiment by directly transferring our model trained on MS-COCO to Flickr30k dataset. For comparison, except for two existing work \cite{liu2017learning,engilberge2018finding} that provide the corresponding results, we additionally re-implement two previous studies \cite{2018VSE++,2018SCAN} based on their open-released code. From Table \ref{tab.4}, it's obvious that our CVSE outperforms all the competitors by a large margin. Moreover, compared to the baseline that only employs instance-level alignment (CVSE$_{wo/consensus}$), CVSE achieves compelling improvements. These results implicate the learned consensus knowledge can be shared between cross-domain heterogeneous data, which leads to significant performance boost.

\setlength{\tabcolsep}{4pt}
\begin{table}[t]
	\setlength{\belowcaptionskip}{+0.15cm}
	\begin{center}
		\caption{Comparison results on cross-dataset generalization from MSCOCO to Flickr30k.}
		\setlength{\abovecaptionskip}{-0.3cm}
		\label{tab.4}
		\centering	
		\resizebox{0.5\textwidth}{1.2cm}{	
			\begin{tabular}{l|ccc|ccc}
				\hline \hline
				\multirow{2}{*}{Approaches} &
				\multicolumn{3}{c}{Text retrieval} & \multicolumn{3}{|c}{Image Retrieval}  \\
				&   R@1 & R@5 & R@10 & R@1 & R@5 & R@10 \\
				\hline
				RRF-Net \cite{liu2017learning}   			& 28.8	& 53.8	& 66.4	& 21.3	& 42.7	& 53.7 \\
				VSE++ 	\cite{2018VSE++}			& 40.5	& 67.3	& 77.7	& 28.4	& 55.4	& 66.6 \\
				LVSE	\cite{engilberge2018finding}			& 46.5	& 72.0	& 82.2	& 34.9	& 62.4  & 73.5 \\
				SCAN    \cite{2018SCAN}			& 49.8	& 77.8	& 86.0	& 38.4	& 65.0	& 74.4 \\
				\hline
				CVSE$_{wo/consensus}$		& 49.1 & 75.5 	& 84.3 	& 36.4 	& 63.7 	& 73.3 \\
				{CVSE}		  		& {\textbf{56.4}}	& {\textbf{83.0}}	& {\textbf{89.0}}	& {\textbf{39.9}}	& {\textbf{68.6}}	& {\textbf{77.2}}  \\		
				
				\hline \hline
			\end{tabular}
		}
	\end{center}
	\vspace{-0.6cm}
\end{table}
\setlength{\tabcolsep}{1.4pt}

\begin{figure}[t]
	\begin{center}
		\includegraphics[width=0.85\linewidth,height=2.7cm]{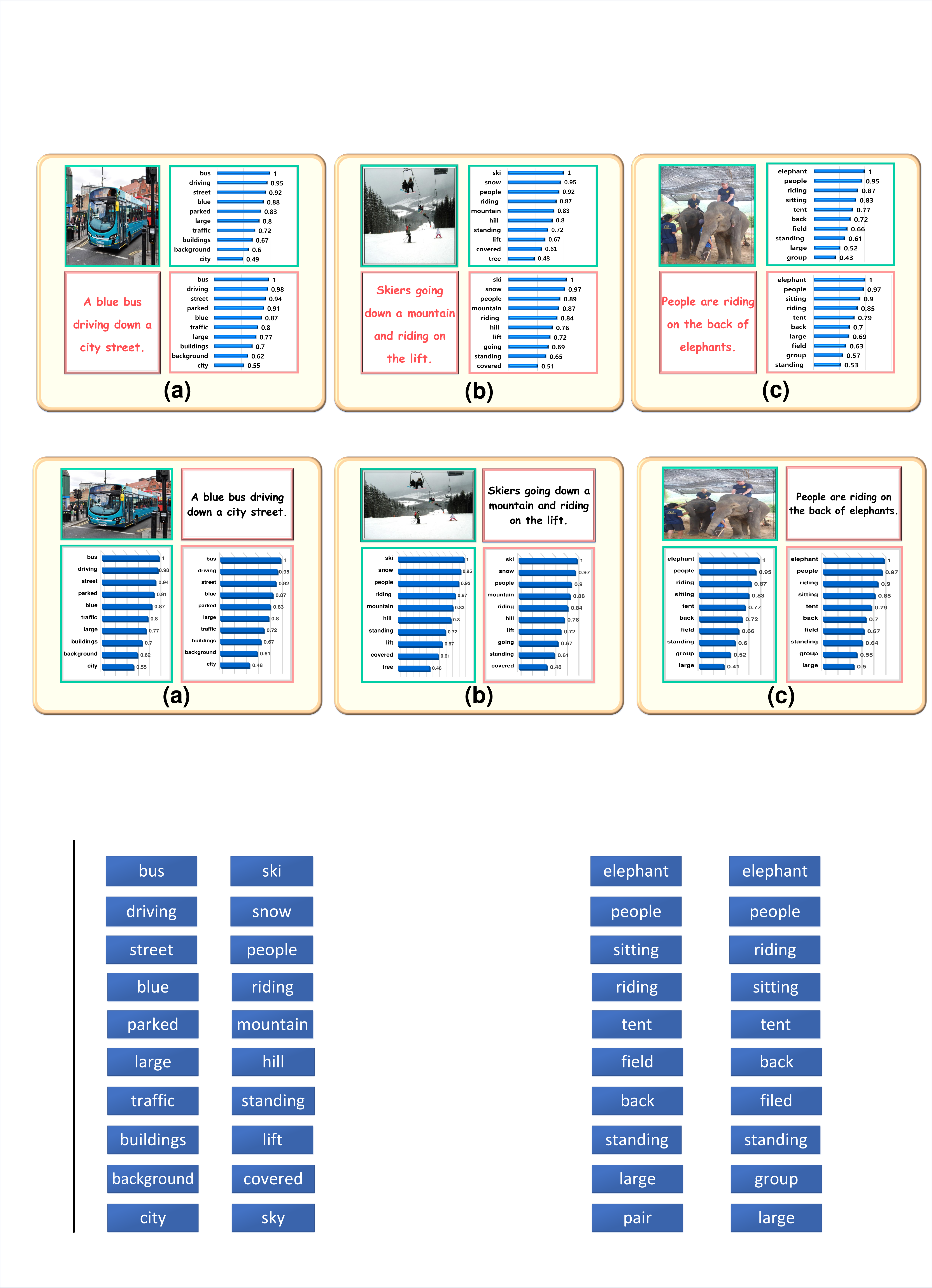}
	\end{center}
	\setlength{\abovecaptionskip}{-0.1cm}
	\caption{The visualization results of predicted scores for top-10 concepts.}
	\label{fig.4}
	\label{fig:long}
	\label{fig:onecol}
	\begin{center}
		\includegraphics[width=0.75\linewidth,height=3.55cm]{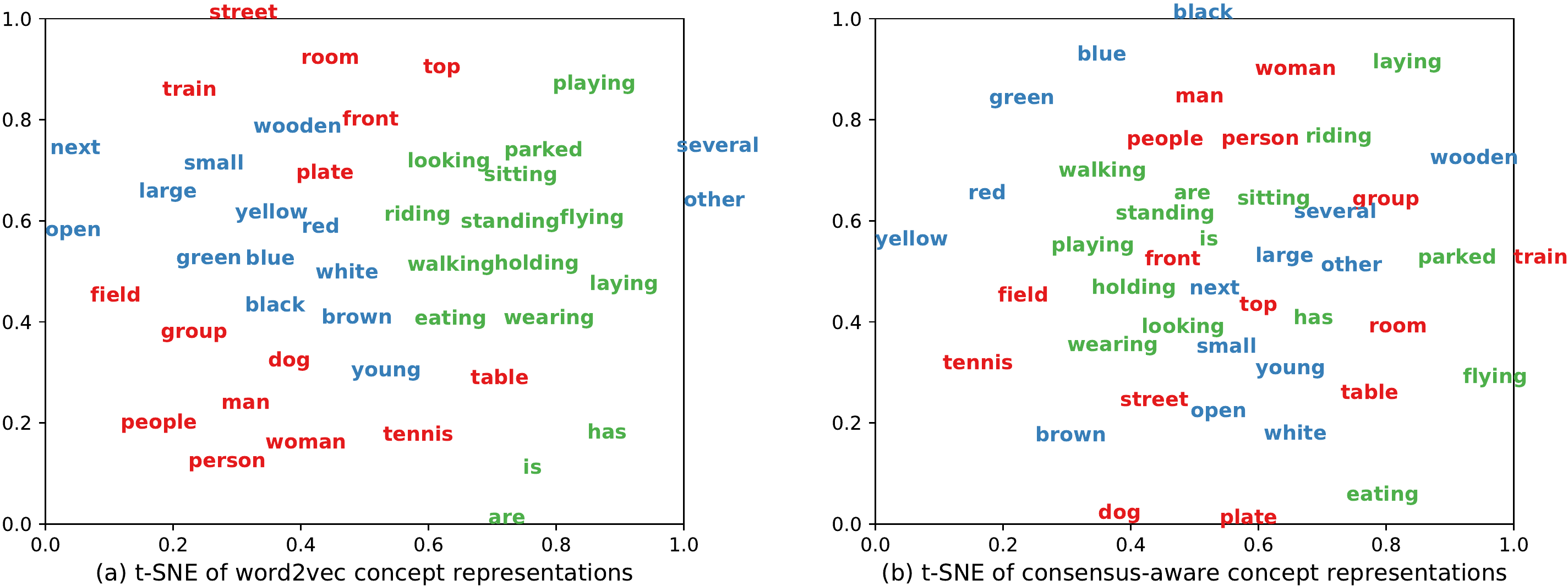}
	\end{center}
	\setlength{\abovecaptionskip}{-0.1cm}
	\caption{The t-SNE results of word2vec based concept representations and  our consensus-aware concepts representations. We randomly select 15 concepts per POS for performing visualization and annotate each POS with the same color.}
	\label{fig.5}
	\label{fig:long}
	\label{fig:onecol}
	\vspace{-0.4cm}
\end{figure}

\subsubsection{The Visualization of Confidence Score for Concepts.}

In Figure \ref{fig.4}, we visualize the confidence score of concepts predicted by our CVSE. It can be seen that the prediction results are considerably reliable. In particular, some informative concepts that are not involved in the image-text pair can even be captured. For example, from Figure \ref{fig.4}(a), the associated concepts of ``\texttt{traffic}'' and ``\texttt{buildings}'' are also pinpointed for enhancing semantic representations.

\subsubsection{The Visualization of Consensus-Aware Concept Representations.}

In Figure \ref{fig.5}, we adopt the t-SNE \cite{maaten2008visualizing} to visualize the CAC representations. In contrast to the word2vec \cite{pennington2014glove} based embedding features, the distribution of our CAC representations is more consistent with our common sense. For instance, the concepts with POS of \textit{Motion}, such as ``\texttt{riding}'', is closely related to the concept of ``\texttt{person}''. Similarly, the concept of ``\texttt{plate}'' is closely associated with ``\texttt{eating}''.  These results further verify the effectiveness of our consensus exploitation module in capturing the semantic associations among the concepts.

\section{Conclusions}

The ambiguous understanding of multi-modal data severely impairs the ability of machine to precisely associate images with texts. In this work, we proposed a Consensus-Aware Visual-Semantic Embedding (CVSE) model that integrates commonsense knowledge into the multi-modal representation learning for visual-semantic embedding. Our main contribution is exploiting the consensus knowledge to simultaneously pinpoint the high-level concepts and generate the unified consensus-aware concept representations for both image and text. We demonstrated the superiority of our CVSE for image-text retrieval by outperforming state-of-the-art models on widely used MSCOCO and Flickr30k datasets.

\section{Acknowledgments}
This work was supported by the Natural Science Foundation of Tianjin under Grant 19JCYBJC16000, and the National Natural Science Foundation of China (NSFC) under Grant 61771329.

\clearpage
%
%
\bibliographystyle{splncs04}

\end{document}